\documentclass[AMA,STIX1COL]{WileyNJD-v2}
\usepackage{moreverb}
\usepackage{algorithm}
\usepackage{graphicx}
\usepackage{multirow}
\usepackage[english]{babel}
\usepackage[T1]{fontenc}
\usepackage{amsmath}
\usepackage{graphicx}
\usepackage{amsfonts}
\usepackage[utf8]{inputenc} 
\usepackage{hyperref}       
\usepackage{url}            
\usepackage{booktabs}       
\usepackage{nicefrac}       
\usepackage{microtype}      
\usepackage{caption}
\usepackage{subcaption}
\usepackage{siunitx}
\usepackage{paralist}
\usepackage{color}

\newcommand\BibTeX{{\rmfamily B\kern-.05em \textsc{i\kern-.025em b}\kern-.08em
T\kern-.1667em\lower.7ex\hbox{E}\kern-.125emX}}

\articletype{Article Type}%

\received{<day> <Month>, <year>}
\revised{<day> <Month>, <year>}
\accepted{<day> <Month>, <year>}
\begin{document}

\title{Dexterous Robotic Manipulation using Deep Reinforcement Learning and Knowledge Transfer for Complex Sparse Reward-based Tasks}

\author[1]{Qiang Wang}

\author[2,3]{Francisco Roldan Sanchez}

\author[1]{Robert McCarthy}

\author[1]{David Cordova Bulens}

\author[2,3]{Kevin McGuinness}

\author[2,3]{Noel O’Connor}

\author[4]{Manuel Wüthrich}

\author[5]{Felix Widmaier}

\author[6]{Stefan Bauer}

\author[1,3]{Stephen J. Redmond}

\authormark{WANG \textsc{et al.}}

\address[1]{\orgdiv{\orgname{University College Dublin}, \country{Ireland}}}
\address[2]{\orgdiv{\orgname{Dublin City University}, \country{Ireland}}}
\address[3]{\orgdiv{\orgname{Insight SFI Research Centre for Data Analytics}, \country{Ireland}}}
\address[4]{\orgdiv{\orgname{Harvard University}, \country{USA}}}
\address[5]{\orgdiv{\orgname{MPI for Intelligent Systems, Tübingen}, \country{Germany}}}
\address[6]{\orgdiv{\orgname{KTH Stockholm}, \country{Sweden}}}

\corres{*S.J. Redmond, UCD School of Electrical and Electronic Engineering, Belfield, Dublin 4, Ireland. \email{stephen.redmond@ucd.ie}}


\abstract[Abstract]{This paper describes a deep reinforcement learning (DRL) approach that won Phase 1 of the Real Robot Challenge (RRC) 2021, and then extends this method to a more difficult manipulation task. The RRC consisted of using a TriFinger robot to manipulate a cube along a specified positional trajectory, but with no requirement for the cube to have any specific orientation. We used a relatively simple reward function, a combination of a goal-based sparse reward and a distance reward, in conjunction with Hindsight Experience Replay (HER) to guide the learning of the DRL agent (Deep Deterministic Policy Gradient (DDPG)). Our approach allowed our agents to acquire dexterous robotic manipulation strategies in simulation. These strategies were then deployed on the real robot and outperformed all other competition submissions, including those using more traditional robotic control techniques, in the final evaluation stage of the RRC. Here we extend this method, by modifying the task of Phase 1 of the RRC to require the robot to maintain the cube in a particular orientation, while the cube is moved along the required positional trajectory. The requirement to also orient the cube makes the agent less able to learn the task through blind exploration due to increased problem complexity. To circumvent this issue, we make novel use of a \textit{Knowledge Transfer} (KT) technique that allows the strategies learned by the agent in the original task (which was agnostic to cube orientation) to be transferred to this task (where orientation matters). KT allowed the agent to learn and perform the extended task in the simulator, which improved the average positional deviation from 0.134 m to 0.02 m, and average orientation deviation from 142$^\circ$ to 76$^\circ$ during evaluation. This KT concept shows good generalization properties and could be applied to any actor-critic learning algorithm.}



\maketitle

\section{Introduction}
Dexterous robotic manipulation is essential in many industrial and domestic settings. Traditional robotic manipulation controllers often rely on solving inverse kinematic equations \cite{manipulation-review}. The goal of this approach is to find the robotic joint angle time course to move the end-effector of a robotic system (arm, gripper, fingers, etc.) to a desired pose \cite{inverse-kin}. Because the solution to this problem is not unique, motion primitives (i.e., a set of pre-computed movements that a robot can take in a given environment) are typically used \cite{motion-primitives,motion-primitives2}. These primitives can each have a defined cost, allowing the robot to avoid non-smooth or other undesirable transitions. However, these techniques have poor generalization ability and require complex control system structures. More precisely, they require significant bespoke tailoring for each novel manipulation task, leading to high implementation times and costs. Moreover, current state-of-the-art in traditional robotic control strategies generally struggle in unstructured tasks, which require high degrees of dexterity. 

Reinforcement learning (RL), a data-driven learning approach in which models are trained by rewarding or punishing an agent that is acting in a particular environment, shows promise \cite{rubiks} to replace traditional robotic control approaches. The goal of this learning paradigm is to maximize the cumulative sum of scalar or vector reward signals received by the agent in response to the actions that it takes \cite{rlbook} so that it learns how to interact appropriately with the environment; that is, it learns how to act in order to maximize the rewards it can receive. However, traditional RL is unable to solve tasks with continuous action and state spaces, due to their high dimensionality. In other words, there are infinite combinations of states and actions in a continuous space, and therefore traditional RL algorithms cannot learn the required mappings between state, action, and reward for these problems. Hence, the field of deep reinforcement learning (DRL) \cite{dqn} has emerged; a new discipline combining deep learning (DL) \cite{dl} and RL, which inherits the capacity to deal with high-dimensional continuous data from DL and decision-making ability from RL. DRL methods have been able to obtain outstanding results in robotic manipulation \cite{drl-review-1,drl-review-2}.  

However, the data inefficiency of DRL is a major barrier to its application in real-world robotics: real robot data collection is time-consuming and expensive. Much DRL research to-date has focused on improving these data-efficiency issues. Due to their generally improved sample complexity, off-policy DRL methods \cite{ddpg,sac} are often preferred to on-policy methods \cite{ppo,trpo}. Model-based DRL methods, which explicitly learn a model of the environment, have been used to further improve sample efficiency \cite{pilco,mbpo,dream}, and have seen success in real robot settings \cite{boadong}. Moreover, offline DRL techniques seek to leverage previously collected data to accelerate learning \cite{offline}, and are able to learn dexterous real-world skills such as opening a drawer \cite{awac}. Imitation learning methods provide the policy with expert demonstrations to learn from \cite{imitation,coarse}, enabling success in real robot tasks, such as peg insertion \cite{leverage imitation}. Finally, simulation-to-real (sim-to-real) transfer methods train a policy quickly and cheaply in simulation before deploying it on the real robot, where learning is significantly slower, and have notably been used to solve a Rubik's cube with a robot hand \cite{rubiks}. To account for simulator modelling errors, and to improve the policies ability to generalize to the real robot, sim-to-real approaches often employ domain randomization \cite{sim2real,domain-randomization} or domain adaptation \cite{meta,adapt} techniques. Domain randomization, which has been particularly effective \cite{rubiks}, randomizes the physics parameters in simulation to learn a robust policy that can adapt to the partially unknown physics of the real system.

The costly nature of real-robot experimentation has limited research related to robotic dexterous manipulation. In light of these issues, the Real Robot Challenge (RRC) \cite{rrc} aims to advance the state-of-the-art in robotic manipulation by providing participants with remote access to a TriFinger robotic platform (see Figure \ref{sim-real}(b)) \cite{trifinger}, allowing for free and easy real-robot experimentation. To further support ease of experimentation, users are also provided with a simulation of this robotic system (see Figure \ref{sim-real}(a)). Full details can be found in the `Protocol' section of the RRC website\footnote{\url{https://real-robot-challenge.com/2021}}. 

This paper aims to extend the task of Phase 1 of the RRC 2021 to a more difficult task of moving a cube along a 3D positional trajectory while also maintaining a desired orientation.  The task of Phase 1 of the RRC 2021 consisted of moving a cube along a defined trajectory. So here we increase the difficulty of this task by requiring the cube orientation to match a target orientation throughout the trajectory. We formulated both tasks as pure RL problems. The movement-related skills of the robot are entirely learned in simulation, hence reducing human involvement compared to traditional real-world learning methods. We chose to use the Deep Deterministic Policy Gradient (DDPG) algorithm, as it was shown to have an excellent performance in handling robotic manipulation tasks with continuous action and observation spaces \cite{ddpg}. We used a reward function composed of goal-based sparse rewards in conjunction with Hindsight Experience Replay (HER) \cite{her} to teach the control policy to move the cube to the desired \textit{xy} coordinates (in the first task, controlling cube position) and then also orientation (in the extended task). Simultaneously, a dense distance-based reward is employed to teach the policy to lift the cube to the desired $z$ coordinate. Finally, we use a novel concept that we call \textit{knowledge transfer} (KT) to facilitate learning during the task requiring cube position and orientation be controlled, which is based on the strategies previously-learned during the task requiring only the cube position be controlled. Compared with other methods used to solve a similar task \cite{gpu-paper,ppo}, our KT method has the advantage of being easier to implement, and can be extended to all actor-critic RL algorithms.

This paper is structured as follows: Section \ref{methods} describes the methodology, including the RRC robotic environment (simulator and real robot) and manipulation tasks, DRL algorithms, and the novel KT approach. Section \ref{results} outlines the results of the conducted experiments, and Section \ref{discussion} discusses our proposed approaches' potential for use in DRL for robotic manipulation tasks. Finally, conclusions are drawn on the performance of the proposed approaches in Section \ref{conclusion}.

\section{Methods} \label{methods}
\subsection{Real Robot Challenge (RRC)}
\begin{figure}[h]
     \centering
     \begin{subfigure}[t]{0.48\textwidth}
         \centering
         \includegraphics[width=\textwidth]{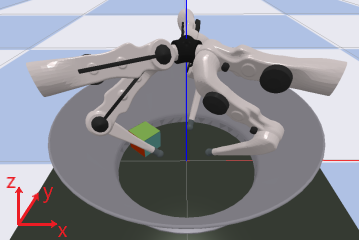}
         \label{tri-sim}
         \captionsetup{font=footnotesize}
         \subcaption{Visualisation of the PyBullet simulation environment for the TriFinger robot used in the RRC 2021.}
     \end{subfigure}
     \hfill
     \begin{subfigure}[t]{0.48\textwidth}
         \centering
         \includegraphics[width=\textwidth]{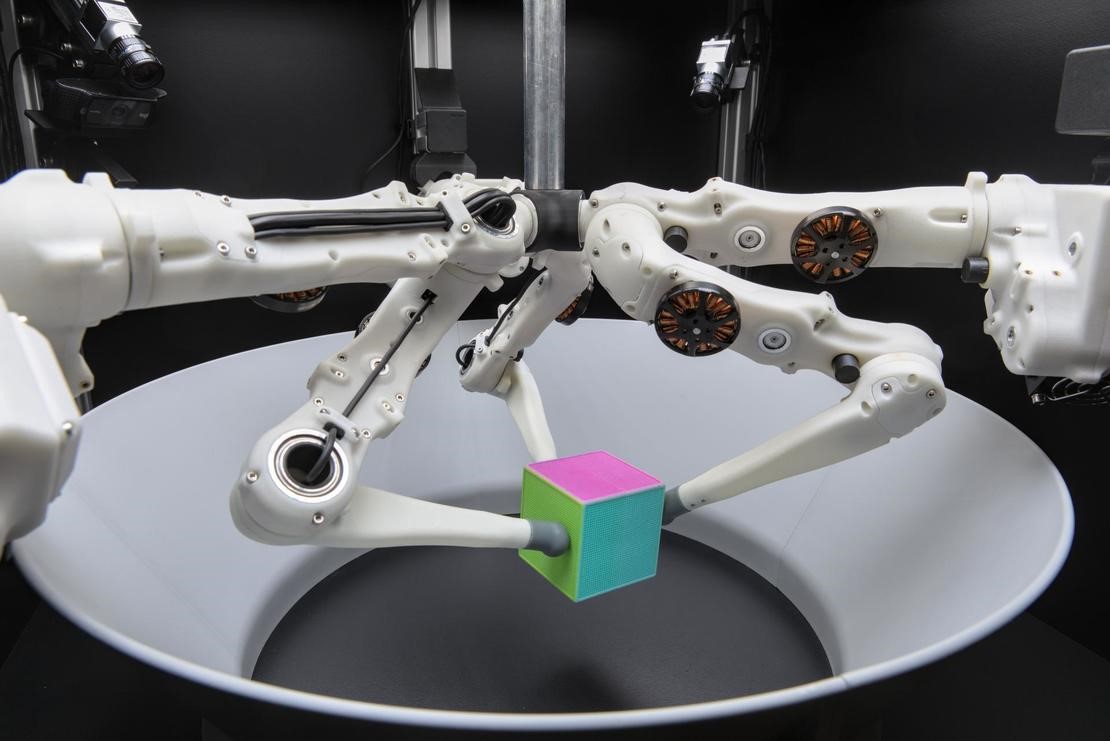}
         \label{tri-real}
         \captionsetup{font=footnotesize}
         \subcaption{The real TriFinger robot used in the RRC 2021. Remote access to this system was possible, allowing competitors from around the world to test the policies learned in simulation on the real robot. Competition results are generated from performance on this real-world system.}
     \end{subfigure}
        \caption{The TriFinger robot in: (a) the PyBullet \cite{pybullet} simulation, and; (b) the real world. Three identical robotic fingers are uniformly placed around the circular arena. The colored cube is the target object that must be moved.}
        \label{sim-real}
\end{figure}

We were provided with remote access to well-maintained TriFinger robotic platforms \cite{trifinger} (see Figure \ref{sim-real}(b)) by the Max Planck Institute for Intelligent Systems (Tübingen/Stuttgart, Germany)\footnote{\url{https://is.mpg.de/}}. To further support ease of experimentation, users are also provided with a simulated version of the robotic setup (see Figure \ref{sim-real}(a)). The 2021 RRC consists of an initial qualifying phase, performed purely in simulation, followed by independent Phases 1 and 2, both performed on the real robot. Full details can be found in the `Protocol' section of the RRC website\footnote{\url{https://real-robot-challenge.com/2021}}.  

\subsubsection{Phase 1 task: Following a 3D position trajectory}
Competition participants are tasked with solving the \textit{`Move Cube on Trajectory'} task. In this task, a cube must be carried along a goal trajectory, which specifies the Cartesian coordinates at which the cube should be positioned at each time-step\footnote{Agents interact with the environment in (discrete) time-steps, which are incremented after the agent takes an action}. The goals are sampled discretely, and the time interval between the two goals in the training phase can be defined manually. The robot should react as soon as the new goal is received to obtain a higher score. For the final evaluation, multiple previously-unseen goal trajectories must be followed. Each evaluation trajectory consists of 120,000 time-steps, with 10 different coordinates that must be visited by the cube. The structure of the trajectory is such that the next goal (of 10 in total) is introduced at each of the following time-steps, $t \in \left\{0,30000,40000,50000,...,110000 \right\}$. The obtained score, $s_{pos}$ is defined by a distance-based weighted criterion: 
\begin{equation}
\label{eq:position critirion}
    s_{pos} = -\dfrac{1}{2} \left(
            \dfrac{\left\|g'_{xy} - g_{xy} \right\|}
            {2d_r} + 
            \dfrac{\left\|g'_{z} - g_{z}\right\|}{d_h} \right)
\end{equation}
\noindent where ${g'}_{xy}$ and ${g'}_{z}$ are the actual $xy$ and $z$ coordinates of the cube, respectively, and $g_{xy}$ and ${g}_{z}$ are the desired $xy$ and $z$ coordinates of the cube. $d_r$ and $d_h$ are constants representing the radius of the arena floor and the maximum height the fingertips can reach. A higher score, averaged across all tested trajectories, implies a better manipulation performance.

Unlike end-to-end pick-and-place manipulation, the goals are continuously changing in this task. Various action strategies, such as pushing, cradling, or pinching, must be dynamically employed to move the cube in response to the changing positional goals. 

\subsubsection{Extended task with increased difficulty: Position and orientation trajectory}
We extended the task described above by introducing the an orientation trajectory goal in addition to the positional goal; this was inspired by the most challenging task in the RRC 2020\footnote{\url{https://real-robot-challenge.com/2020}}, where the cube must not only be moved to the desired position but also aligned with the desired orientation. The task in 2020 is a single end-to-end movement, rather than trajectory. Hence, we integrated position, orientation, and trajectory together to create a more challenging task that we named \textit{Move Cube on Trajectory Pro}. To evaluate performance during this task we used the criterion used in RRC 2020 (c.f., Eq. (\ref{eq:position orientation critirion}))
\begin{equation}
\label{eq:position orientation critirion}
    s_{pos+ori} = -{\left\| \left(R\left(g'_{o}\right)\right)^{-1}    R\left(g_{o}\right) \right\|} + s_{pos}
\end{equation}
\noindent where the ${g'}_{o}$ and ${g}_{o}$ are the achieved orientation and the desired orientation, respectively, described using quaternions. \textit{R} is the rotation matrix derived from the quaternions representation and $^{-1}$ is the inverse matrix operator. Note, $s_{pos}$ from Eq. (\ref{eq:position critirion}) is incorporated in the calculation of $s_{pos+ori}$ to evaluate the cube's position. 


\subsection{Simulated environment}
\subsubsection{Actions and observations} 
Pure torque control of the robot arms is employed with an update frequency of 20 Hz (i.e., each time-step in the environment is 0.05 seconds). The robot has three arms, with three motorized joints in each arm; thus, the action space is 9-dimensional (and continuous). Observations include: (i) robot joint positions, velocities, and torques; (ii) cube's current pose (i.e., its position and orientation, which in simulation is read directly from the simulator with no measurement error, and for the real world arena is estimated using provided computer vision object detection and segmentation methods), along with the difference between the poses at current and previous time-steps; and (iii) the current goal pose. In total, the observation space has 44 dimensions for the \textit{Move Cube on Trajectory} task, and 48 for the \textit{Move Cube on Trajectory Pro}.

In the \textit{Move Cube on Trajectory} task, there are three target variables (three Cartesian position coordinates). In the \textit{Move Cube on Trajectory Pro} task, there are seven target variables, representing 3D position and an additional four quaternion variables representing 3D orientation. 

\subsubsection{Episodes} 
In each simulated training and testing episode, the robot begins in its default position. The simulator instantiates the cube and the arena environment, with the cube's initial position randomly sampled from the arena floor. Each episode lasts 90 time-steps, and the goal trajectory contains three desired goals, which are randomly sampled from the arena 3D space. The intervals between goals are same; i.e., the goal changes every 30 time-steps.

\subsubsection{Domain randomization}
To help the learned policy generalize from a potentially inaccurate simulation to the real environment, we used some basic domain randomization (DR) techniques\footnote{Our domain randomization implementation is based on the benchmark code from the 2020 RRC \cite{dr-code}.} (i.e., physics randomization). This includes uniformly sampling, from a specified range, parameters of the simulation physics (e.g., robot mass, restitution, damping, friction; see our code for more details\footnote{\url{https://github.com/RobertMcCarthy97/rrc_phase1}}) and cube properties (mass and width) for each episode. To account for noisy real robot actuation and observations, uncorrelated noise is added to actions and observations during simulated episodes.

DR is only applied in the \textit{Move Cube on Trajectory} task, since it requires sim-to-real transfer. In our final approach, we train the agent initially in the simulated non-DR environment for 300 epochs to allow it to learn the optimal policy quickly and easily. Afterward, the agent is tuned in the simulated DR environment for 100 epochs to enhance its robustness, before deploying to the real TriFinger robot. 

\subsection{Learning algorithm}
The goal-based nature of the \textit{Move Cube on Trajectory} and \textit{Move Cube on Trajectory Pro} tasks makes HER a good choice of learning algorithm; HER has excelled in similar goal-based robotic tasks \cite{her} and obviates the need for complex reward engineering. As such, we combine DDPG and HER to make our RL algorithm\footnote{Our DDPG + HER implementation is taken from \url{https://github.com/TianhongDai/hindsight-experience-replay}, and uses hyperparameters largely based on \cite{fetch results}.}. However, in our early experiments we observed that the standard DDPG plus HER learning algorithm was slow in learning to lift the cube. To resolve this issue, we slightly altered the HER process and incorporated an additional dense reward which encourages cube-lifting behaviors. We describe these elements of the learning algorithm below.

\subsubsection{Goal-based Reinforcement Learning}
We frame the RRC robotic environment as a Markov decision process (MDP), defined by the tuple \((\mathcal{S},\mathcal{A},\mathcal{G},p,r,\gamma,\rho_0)\). $\mathcal{S}$, $\mathcal{A}$, and $\mathcal{G}$ are the state, action and goal spaces, respectively. The state transition distribution is denoted as \(p(s'|s,a)\), the initial state distribution as \(\rho_0(s)\), and the reward function as \(r(s,g)\). \(\gamma\in(0,1)\) discounts future rewards. The goal of the RL agent is to find the optimal policy $\pi^{*}$ that maximizes the expected sum of discounted rewards in this MDP over all time, \textit{t}: \(\pi^{*} = \operatorname*{argmax}_\pi \mathbb{E}_{\pi} [\sum_{t=0}^{\infty} \gamma^{t} r(s_t,g_t)]\).

\subsubsection{Deep Deterministic Policy Gradients (DDPG)}
DDPG \cite{ddpg} is an off-policy RL algorithm which, in the goal-based RL setting, maintains the following neural networks: a policy (actor) \(\pi:\mathcal{S} \times \mathcal{G} \rightarrow \mathcal{A}\), and an action-value function (critic) $Q:\mathcal{S} \times \mathcal{G} \times \mathcal{A} \rightarrow \mathbb{R} $. The critic is trained to minimize the loss \(\mathcal{L}_c = \mathbb{E}(Q(s_{t},g_{t},a_{t})-y_{t})^2\), where $y_{t}=r_{t} + \gamma Q(s_{t+1},g_{t+1}, \pi(s_{t+1},g_{t+1}))$. To stabilize the critic's training, the targets $y_{t}$ are produced using slowly updated polyak-averaged versions of the main networks. The actor is trained to minimise the loss: $\mathcal{L}_a = -\mathbb{E}_s[ Q(s,g,\pi(s,g))]$, where gradients are computed by backpropagating through the combined critic and actor networks. For these updates, the transition tuples $(s_t,g_t,a_t,r_t,s_{t+1},g_{t+1})$ are sampled from a replay buffer which stores previously collected experiences (i.e., off-policy data).

\subsubsection{Hindsight Experience Replay (HER)}
HER \cite{her} can be used with any off-policy RL algorithm  in goal-based tasks, and is most effective when the reward function is sparse and binary (e.g. equation (\ref{eq:xyonly})). To improve learning in the sparse reward setting, HER employs a simple trick when sampling previously collected transitions for policy updates: a proportion of sampled transitions have their goal $g$ altered to $g’$, where $g’$ is a goal achieved later in the episode. The rewards of these altered transitions are then recalculated with respect to $g’$, leaving the altered transition tuples as $(s_{t},g_{t}',a_{t},r_t',s_{t+1},g_{t+1}')$. Even if the original episode was unsuccessful, these altered transitions will teach the agent to achieve $g'$, thus accelerating its acquisition of skills.

\subsubsection{Rewards and HER}
In our approach, the reward function that guides the RL agent's learning consists of three components: (i) a sparse reward based on the the cube's $xy$ coordinates, termed $r_{xy}$; (ii) a dense reward based on the cube's $z$ coordinate, termed $r_{z}$ (the coordinate frame can be seen in Figure \ref{sim-real}(a)); and (iii) a sparse reward based on the cube's 3D orientation (used in the \textit{Move Cube on Trajectory Pro} task only).

The sparse $xy$ reward, $r_{xy}$, is calculated as:

\begin{equation}
\label{eq:xyonly}
    r_{xy} =  \begin{cases} 
      0 & \textrm{if} \quad \left\| {g'}_{xy} - g_{xy} \right\|\leq 2 \textrm{ cm} \\
     -1 & \textrm{otherwise}
   \end{cases}
\end{equation}

\noindent where ${g'}_{xy}$ are the $xy$ coordinates of the \textit{achieved} goal (the actual $xy$ coordinates of the cube), and $g_{xy}$ are the $xy$ coordinates of the \textit{desired} goal.

The dense $z$ coordinate reward, $r_{z}$, is defined as:
\begin{equation}
\label{eq:z}
  r_{z} =  \begin{cases} 
      - a \left\| z_{cube} - z_{goal}\right\|  & \textrm{if} \quad z_{cube} <  z_{goal} \\
      \\
      \dfrac{-a}{2}\left\| z_{cube} - z_{goal}\right\|  & \textrm{if } \quad z_{cube} > z_{goal}
   \end{cases}
\end{equation}
\noindent where $z_{cube}$ and $z_{goal}$ are the $z$ coordinates of the cube and goal, respectively, and $a$ is a parameter which weights $r_{z}$ relative to $r_{xy}$; we use $a=20$.

The sparse reward for the orientation is defined as:
\begin{equation}
\label{eq:orientation}
    r_{ori} =  \begin{cases} 
      0 & \textrm{if} \quad \left\| \left(R\left(g'_{o}\right)\right)^{-1}
    R\left(g_{o}\right) \right\|\leq 0.384 \textrm{ rad (i.e., $22^{\circ}$)} \\
     -1 & \textrm{otherwise}
   \end{cases}
\end{equation}
\noindent where the ${g'}_{o}$ and ${g}_{o}$ are the \textit{achieved} orientation and the \textit{desired} orientation represented as quaternions. \textit{R} indicates rotation matrix calculation and $^{-1}$ is the matrix inverse operator. We set the position ($2$ cm) and orientation ($22^{\circ}$) thresholds as per Andrychowicz \textit{et al.} \cite{openai}.

Another pure distance-based reward is also used as a comparator for our position-based reward scheme. This reward is popular in the field because of its clear geometric meaning and its computational efficiency:
\begin{equation}
\label{eq:pure-dis}
    r_{dis} = -{\left\| g'_{xyz} - g_{xyz}\right\|}
\end{equation}
\noindent where ${g'}_{xyz}$ and  ${g}_{xyz}$ are the \textit{achieved} $xyz$ coordinates and the \textit{desired} $xyz$ coordinates of the cube.

For the position-only \textit{Move Cube on Trajectory} task, we only apply HER to the $xy$ coordinates of the goal; i.e., the $xy$ coordinates of the goal can be altered in hindsight, but the $z$ coordinate remains unchanged. Thus, our HER altered goals are: $\hat{g} = (g'_{xy},g_z)$, meaning only $r_{xy}$ is recalculated after HER is applied to a transition sampled during policy updates. This reward system is motivated by the following:

\begin{enumerate}
\item Using $r_{xy}$ with HER allows the agent to learn to push the cube around in the early stages of training, even if it cannot yet lift the cube to reach the $z$ coordinate of the goal. As the agent learns to push the cube around in the $xy$ plane of the arena floor, it can then more easily stumble upon actions which lift it. Importantly, the approach of using $r_{xy}$ with HER requires no bespoke reward engineering.

\item $r_{z}$ aims to explicitly teach the agent to lift the cube by encouraging minimization of the vertical distance between the cube and the goal. It is less punishing when the cube is above the goal, serving to further encourage lifting behaviours.

\item In the early stages of training, the cube mostly remains on the floor. During these stages, most $g'$ sampled by HER will be on the floor. Thus, applying HER to $r_{z}$ could often lead to the agent being punished for briefly lifting the cube. Since we only apply HER to the $xy$ coordinates of the goal, our HER altered goals, $\hat{g}$, maintain their original $z$ height. This leaves more room for the agent to be rewarded by $r_{z}$ for any cube lifting it performs.
\end{enumerate}

In the \textit{Move Cube on Trajectory Pro} task, the reward of the positional components is same as above. HER is applied on the orientation and thus the HER altered goals are $\hat{g} = (g'_{xy},g'_{ori},g_z)$. Finally all items are summed to form the final reward, $r=r_{xy}+r_{z}+r_{ori}$ 

\subsubsection{Goal trajectories}
In each episode, the agent is faced with multiple goals; it must move the cube from one goal to the next along a given trajectory. To ensure the HER process remains meaningful in these multi-goal episodes, we only sample future achieved goals, $g'$, (to replace $g$) from the period of time in which $g$ was active.

In our implementation, the agent is unaware that it is dealing with trajectories: when updating the policy with transitions $(s_{t},g_{t},a_{t},r_{t},s_{t+1},g_{t+1})$ we always set $g_{t+1} = g_{t}$, even if in reality $g_{t+1}$ was different\footnote{Interestingly, we found that exposing the agent (during updates) to transitions in which $g_{t+1} \neq g_{t}$ hurt performance significantly, perhaps due to the extra uncertainty this introduces to the DDPG action-value estimates.}. Thus, the policy focuses solely on achieving the current active goal and is unconcerned with any future changes in the active goal.

\subsubsection{Exploration vs exploitation}
\begin{figure}[h]
    \centering
    \includegraphics[width=0.8\textwidth]{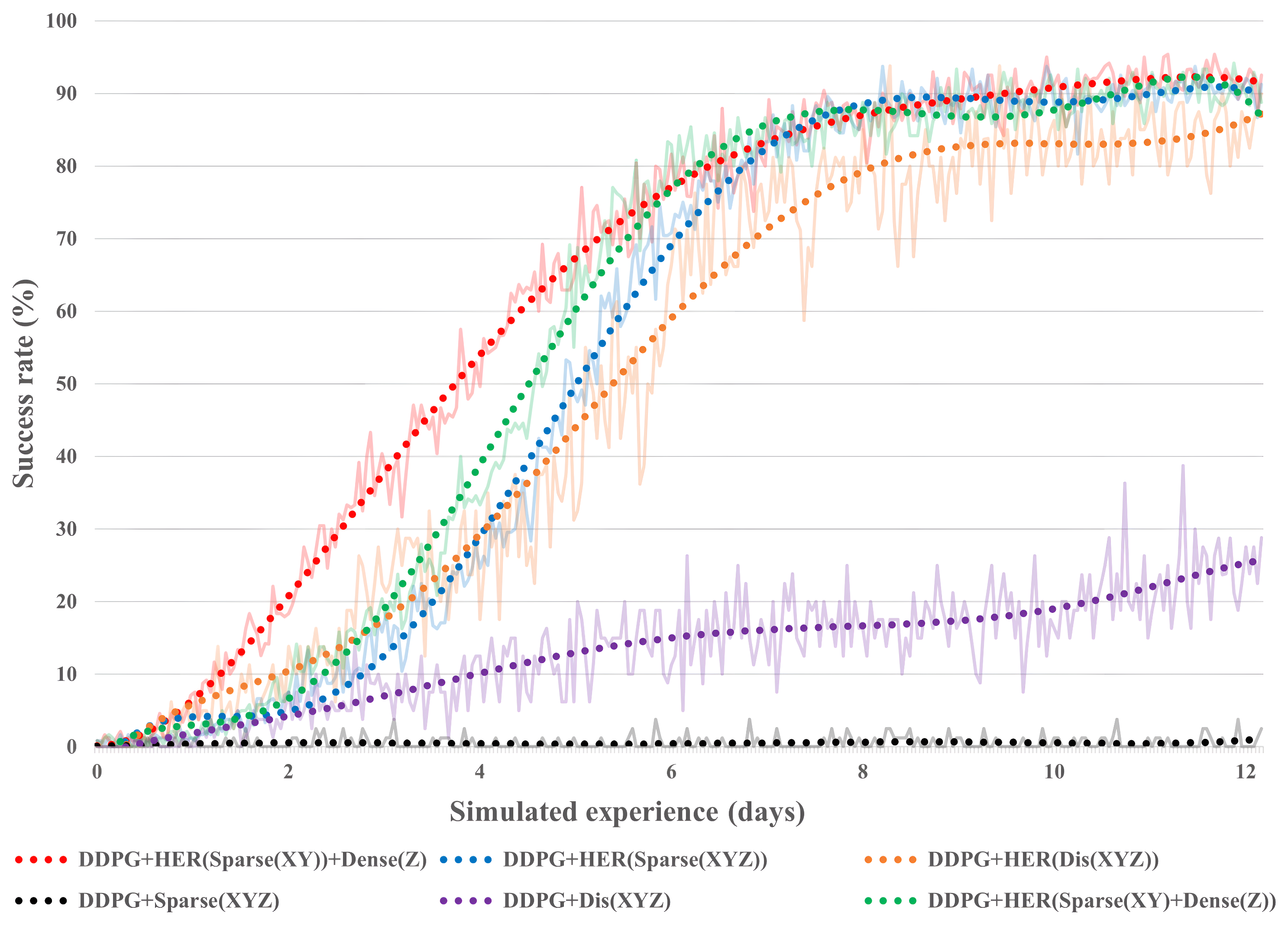}
    \caption{Success rate vs experience collected during simulated training (1 day in simulator, $\approx$ 1.7 million environment steps). We took the average over 3 randomly-selected seeds for this illustration. The solid dotted lines are the trend lines and the faded lines are the raw data. An episode is deemed successful if, when complete, the final goal of the trajectory has been achieved. We compare training with: (i) Our final method where HER is applied to $r_{xy}$ but not to $r_z$ (red); (ii) HER applied to both $r_{xy}$ and $r_z$ (green); (iii) HER applied to a standard sparse reward where $xyz$ are calculated together (blue); (iv) HER applied to pure distance reward, $r_{dis}$, where $xyz$ are calculated together (orange); and (v) Pure distance reward, $r_{dis}$, without HER (purple); (vi) Standard sparse reward without HER (black).
    }
    \label{sim train}
\end{figure}

We derive our DDPG-HER hyperparameters from Plappert \textit{et al.} \cite{fetch results}, who use a highly exploratory policy when collecting data in the environment: with 30\% probability a random action is sampled (uniformly) from the action space, and when policy actions are chosen, Gaussian noise is applied. This is beneficial for exploration in the early stages of training, however, it can be limiting in the later stages when the policy must be fine-tuned; we found that the exploratory policy repeatedly drops the cube due to the randomly sampled actions and the injected action noise. To resolve this issue, rather than slowly reducing the level of exploration each epoch (which would require a degree of hyperparameter tuning), we make efficient use of evaluation episodes (which are performed by the standard exploitation policy) by adding them to the replay buffer. Thus, 90\% of rollouts added to the buffer are collected with the exploratory policy, and the remaining 10\% with the exploitation policy. This addition was sufficient to boost final success rates in simulation from 70-80\% to >90\%  (where "success rate" is equivalent to that seen in Figure \ref{sim train}).  

\subsection{Knowledge transfer}
The signal from our computationally-simple reward rarely guides the agent to focus on playing with the cube in the early training stage; hence, the agent's learning relies heavily on pure exploration. However, its arbitrary exploration is unlikely to explore actions that change the cube's orientation. Unlike the position-only \textit{Move Cube on Trajectory} task, the orientation exploration in the \textit{Move Cube on Trajectory Pro} task is comparatively more difficult, requiring the cube to be both lifted and rotated simultaneously. To this end, we introduce our KT approach, which transfers a trained agent's knowledge (\textit{teacher}) to another untrained agent (\textit{student}), or uses the \textit{teacher}'s knowledge to assist the \textit{student} with learning. In our case, we train a \textit{teacher} from the position-only \textit{Move Cube on Trajectory} task, and train \textit{students} in the \textit{Move Cube on Trajectory Pro} task. The \textit{teacher's} transferred knowledge increases the likelihood of the \textit{student} discovering useful actions to control the orientation of the cube through exploring in the early stages of training, instead of exploring all possible actions at random, as the \textit{student} has been transferred the knowledge required to control the cube's position; in this way, we expect the agent to learn how to control the cube's orientation more quickly. We use two strategies to ensure the \textit{teacher's} guidance is not too biased toward position-based exploration: 1) increasing the action noise in the early stages of exploration; 2) choose a \textit{teacher} that has weaker performance in the \textit{Move Cube on Trajectory} task (i.e., success rate of only 80\%). 

We have the following three implementations of KT. We name them differently for ease of explanation later.

\begin{enumerate}

\item \textit{\textbf{ACTOR-CRITIC}}: Initialise the \textit{student's} actor and critic\footnote{The \textit{actor} is typically a policy function, which outputs actions to interact with the environment. The \textit{critic} uses approximate architecture and simulations to learn a value function, which is then used to update the \textit{actors'} policy parameters \cite{ac}.} network weights by loading the \textit{teacher's} trained weights. 

\item \textit{\textbf{ACTOR}}: Initialise the \textit{student}'s actor network weights by loading the \textit{teacher's} trained weights and randomly initialize the \textit{student's} critic. 

\item \textit{\textbf{COLLECT}}: The \textit{student} does not inherit any network weights from the \textit{teacher}. The \textit{teacher} only helps the \textit{student} to collect experience in the early stages of training. Both the \textit{student's} actor and critic networks are randomly initialized. As the \textit{teacher's} knowledge is not updated in the new task, the collected experience lacks diversity. Hence, we force the \textit{teacher's} participation to decay as training progresses, allowing the \textit{students} to engage in exploration further, making the experience more beneficial to the \textit{student}.

\item \textit{\textbf{SCRATCH}}: The attempt to solve the \textit{Move Cube on Trajectory Pro} task, `from scratch', without using KT. This is used for comparing KT approaches.

\end{enumerate}

\subsection{Train, test, and evaluate}
The structure of the neural networks and hyperparameters that we used for training are shown in Appendix \ref{neural-network-params}. All training and testing were performed on the Sonic high performance cluster at University College Dublin, Ireland. For our training, jobs are randomly assigned to Dell machines configured with R640 2x Intel Xeon Gold 6152 (2.1 GHz, 22 cores) or C6200 V2 2x Intel E5-2660 v2 (2.2 GHz, 10 cores). In each training run, eight RL agents are assigned to eight processors and run in parallel. At the end of each training step, the neural network weights of eight agents are synchronized by averaging. The experiences collected by agents are stored in their respective experience replay buffers and will not be shared. The global success rate and reward are calculated by averaging the local values from eight agents. 

\begin{itemize}
    \item \textit{Training}: For the \textit{Move Cube on Trajectory} task, each job runs for 300 epochs, overall interacting with the simulated environment for 21.6 million time-steps (2.7 million time-steps for each RL agent). The neural network of each agent is updated 15,000 times and synchronization is performed after each update. For the \textit{Move Cube on Trajectory Pro} task, the number of training epochs is increased to 500, and other training parameters increased proportionally.
    

    \item \textit{Testing}: Testing is performed after each training epoch. Each agent runs for 900 time-steps in the simulator. Experience collected from testing is stored in the replay buffer. The neural network losses, success rates and rewards of eight agents are averaged and saved. 

    \item \textit{Real robot evaluation of \textit{Move Cube on Trajectory}}: For the \textit{Move Cube on Trajectory} task performed on the real TriFinger robot, we adopt the evaluation criteria set by the RRC 2021. For each evaluation episode, a random goal trajectory is sampled. Each test run keeps 120,000 time-steps, with 10 different goals (first goal at first time-step, then 30,000 time-steps to the second goal, and 10,000 time-steps for each of the remaining 8 goals after that). The domain gap between simulation and reality was significant, and generally led to inferior scores on the real robot. Policies often struggled to gain control of the real cube which appeared to slip more easily from the fingers than in simulation. Additionally, on the real robot, policies could become stuck with a fingertip pressing the cube into the wall. As a heuristic solution to this issue, we assumed the policy was stuck whenever the cube had not reached the goal's $xy$ coordinates for 50 consecutive time-steps; then uniformly sampled random actions were taken for 7 time-steps in an attempt to free the policy from its stuck state.
    
\end{itemize}

\section{Results} \label{results}
\begin{figure}[t]
     \centering
     \begin{subfigure}[b]{0.3\textwidth}
         \centering
         \includegraphics[width=\textwidth]{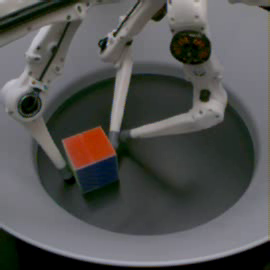}
         \captionsetup{font=footnotesize}
         \subcaption{Pushing}
     \end{subfigure}
     \hfill
     \begin{subfigure}[b]{0.3\textwidth}
         \centering
         \includegraphics[width=\textwidth]{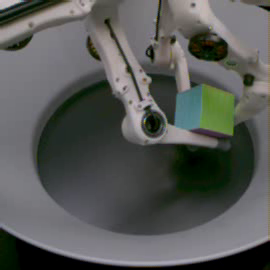}
         \captionsetup{font=footnotesize}
         \subcaption{Cradling}
     \end{subfigure}
     \hfill
     \begin{subfigure}[b]{0.3\textwidth}
         \centering
         \includegraphics[width=\textwidth]{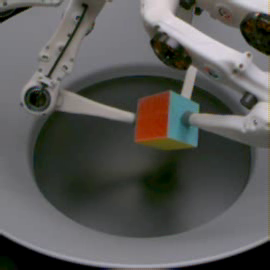}
         \captionsetup{font=footnotesize}
         \subcaption{Pinching}
     \end{subfigure}
        \caption{The various manipulation strategies learned by the agent: (a) pushing; (b) cradling; (c) pinching.}
        \label{fig:three graphs}
\end{figure}

\subsection{RRC 2021 Phase 1}

\subsubsection{Learning outcomes of RL agents}
Our method is highly effective in simulation. The algorithm can learn from scratch to proficiently grasp the cube and lift it along goal trajectories. Figure \ref{sim train} compares the training performance of our final algorithm (red curve) to that of other combinations\footnote{These runs did not use domain randomization. Generally we trained from scratch in standard simulation before fine-tuning in a domain-randomized simulation.}. Our algorithm appears to converge stably and more quickly than either HER applied to a standard sparse reward or HER applied to both $r_{xy}$ and $r_{z}$. Furthermore, HER plays a crucial role in the success of our algorithm. The pure DDPG agent finds it difficult to learn from sparse rewards without HER, and the convergence speed is extremely slow for an agent guided by this traditional distance-based reward. 

Throughout different training runs, our policies learned several different manipulation strategies, the most distinct of which included: `\textit{pushing}' the cube on the arena floor; (ii) `\textit{cradling}' the cube with all three of its forearms (not fingertips); and (iii) `\textit{pinching}' the cube with two fingertips and supporting it with the third (see Figure \ref{fig:three graphs}). 

Our final policies transferred to the real robot with reasonable success. Table \ref{scores} displays the self-reported scores of our best \textit{pinching} and \textit{cradling} policies under RRC Phase 1 evaluation conditions. As a baseline comparison, we trained a simple `\textit{pushing}' policy which ignores the height component of the goal and simply learns to push the cube along the floor to the goal's $xy$ coordinates. The \textit{pinching} policy performed best on the real robot, and is capable of carrying the cube along goal trajectories for extended periods of time, and of recovering the cube when it is dropped. This policy was submitted for the official RRC Phase 1 final evaluation round and obtained the winning score\cite{rrc-win} (see \footnote{\url{https://real-robot-challenge.com/2021\#nav-results}}, username `thriftysnipe').


\begin{table}[]
  \caption{Self-reported (i.e., not reported by competition organizers) evaluation scores of our learned \textit{pushing}, \textit{cradling}, and \textit{pinching} policies when deployed on the simulated and real robots (mean $\pm$ standard deviation (SD) score over 10 episodes). Scores are based on the cumulative position error of the cube during an episode: $ s_{pos} = -\frac{1}{2}\sum_{t=0}^{n} \left(\frac{||\textbf{e}_{xy}^{t}||}{d_{r}} + \frac{|e_{z}^{t}|}{d_{h}}\right) $, where $\textbf{e}^{t} = \left(e_x^t,e_y^t,e_z^t\right)$ is the error between the cube and goal position at time-step $t$, $d_{r}$ the arena floor radius, and $d_h$ the range on the z-axis.}
  \vspace{1.5mm}
  \label{scores}
  \centering
\begin{tabular}{l|rlr|rlr|rlr}
\hline
           & \multicolumn{3}{c|}{Pushing} & \multicolumn{3}{c|}{Cradling} & \multicolumn{3}{c}{Pinching} \\ \hline
Simulation & -20,399    & $\pm$    & 3,799    & -6,349      & $\pm$    & 1,039    & \textbf{-6,198}  & $\pm$ & 1,840 \\ \hline
Real robot & -22,137    & $\pm$    & 3,671    & -14,207     & $\pm$    & 2,160    & \textbf{-11,489} & $\pm$ & 3,790 \\ \hline
\end{tabular}
\end{table}

\subsubsection{Agents trained from different random seeds behave differently}
We verified that discrepancies exist between agents' performances when trained from different random seeds (see Table \ref{real and sim eval different seeds}) \cite{rl-matters}; for detailed experimental results before averaging, please see Appendix \ref{eval-raw-data-ndr}. In our case, the random seed is used to randomly reset the environment and randomize actions so that agents can explore in the training phase. Interestingly, the best performing agent (Seed 0 in Table \ref{real and sim eval different seeds}) in the simulator had a relatively poor performance on the real robot. This agent overfits to the simulator environment, and hence the acquired policy has low generalization capability. The agent train under (random) Seed 200 (see Table \ref{real and sim eval different seeds}) did not obtain the optimal policy and performed the worst on the real robot.

\begin{table*}[]
\renewcommand{\arraystretch}{1.2}
\caption{Evaluated score of RL agents trained under three different random seeds. The RRC organizing institute has a small number of real robots available for remote use, with a small but expected variation in the physical parameters of each arena and the performance of each robot. The evaluations are performed on three of these robots, chosen at random, named Roboch1, Roboch5 and Roboch6, to increase the generalizability of the results. Each evaluation is repeated 15 times (15 goal trajectories) with 120,000 time-steps in each repeat. The results from the simulator are also shown in the Sim subcolumns. The raw results before averaging are shown in Appendix \ref{eval-raw-data-ndr}.}
  \label{real and sim eval different seeds}
  \centering
\resizebox{\textwidth}{!}{
\large
\begin{tabular}{c|cccc|cccc|cccc|cc}
\hline
\multicolumn{1}{l|}{} &
  \multicolumn{4}{c|}{\textbf{Seed 0}} &
  \multicolumn{4}{c|}{\textbf{Seed 123}} &
  \multicolumn{4}{c|}{\textbf{Seed 200}} &
  \multicolumn{2}{c}{\textbf{Average over seeds}} \\ \hline
\textbf{\begin{tabular}[c]{@{}c@{}}Robot\\ (Real/Sim)\end{tabular}} &
  \multicolumn{1}{c|}{Robch1} &
  \multicolumn{1}{c|}{Robch5} &
  \multicolumn{1}{c|}{Robch6} &
  Sim &
  \multicolumn{1}{c|}{Robch1} &
  \multicolumn{1}{c|}{Robch5} &
  \multicolumn{1}{c|}{Robch6} &
  Sim &
  \multicolumn{1}{c|}{Robch1} &
  \multicolumn{1}{c|}{Robch5} &
  \multicolumn{1}{c|}{Robch6} &
  Sim &
  \multicolumn{1}{c|}{Real robots} &
  Sim \\ \hline
\textbf{Reward} &
  \multicolumn{1}{c|}{-12175} &
  \multicolumn{1}{c|}{-11300} &
  \multicolumn{1}{c|}{-9913} &
  \multirow{2}{*}{\textbf{-5447}} &
  \multicolumn{1}{c|}{-9308} &
  \multicolumn{1}{c|}{-8926} &
  \multicolumn{1}{c|}{-10017} &
  \multirow{2}{*}{-6376} &
  \multicolumn{1}{c|}{-16266} &
  \multicolumn{1}{c|}{-13120} &
  \multicolumn{1}{c|}{-15750} &
  \multirow{2}{*}{-7277} &
  \multicolumn{1}{c|}{\multirow{2}{*}{\textbf{-11864±3181}}} &
  \multirow{2}{*}{\textbf{-6367±920}} \\ \cline{1-4} \cline{6-8} \cline{10-12}
\textbf{Avg} &
  \multicolumn{3}{c|}{-11129} &
   &
  \multicolumn{3}{c|}{\textbf{-9417}} &
   &
  \multicolumn{3}{c|}{-15045} &
   &
  \multicolumn{1}{c|}{} &
   \\ \hline
\end{tabular}}

\end{table*}

\begin{figure}[h]
     \centering
        \includegraphics[width=0.8\textwidth]{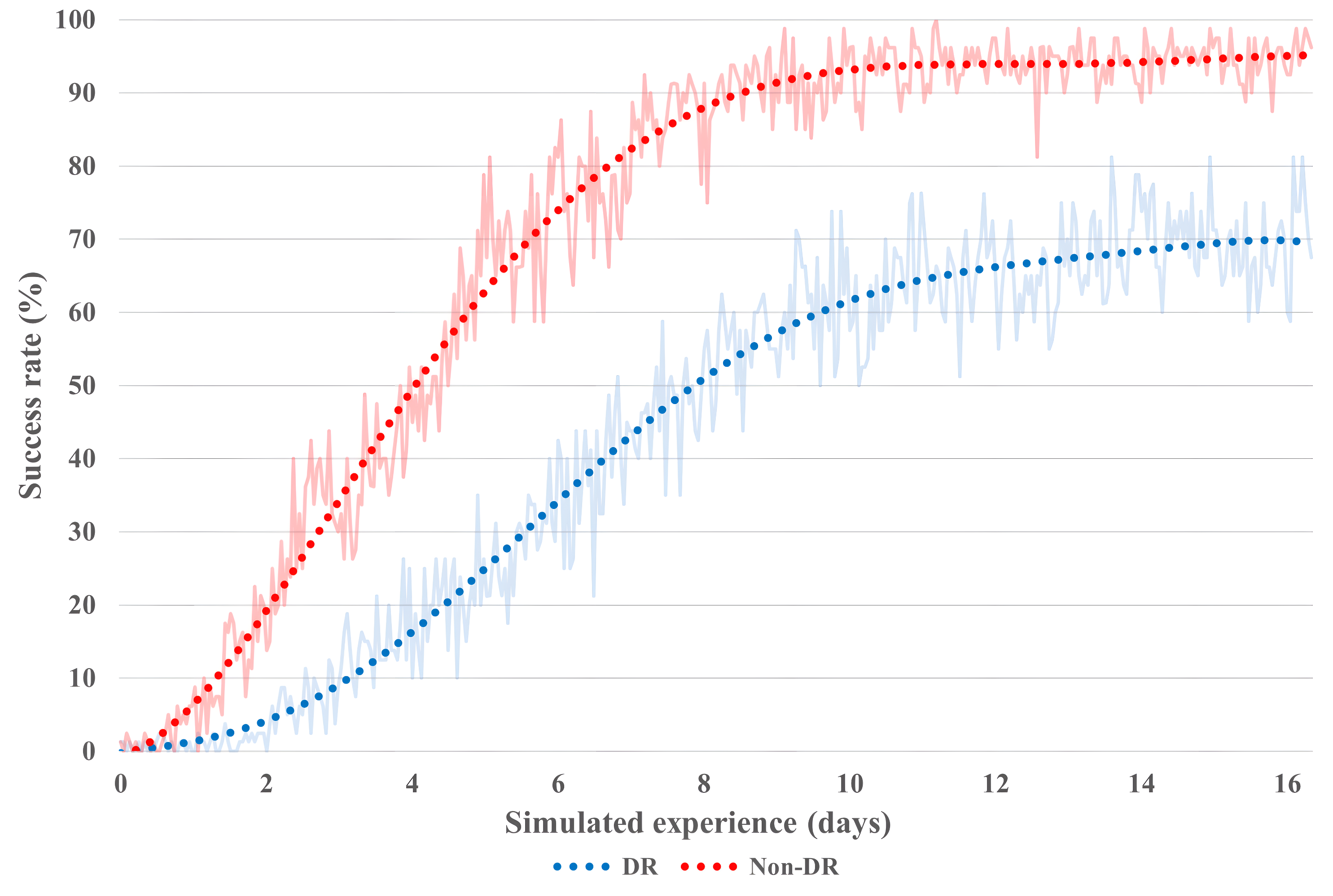}
        \caption{Comparison between agents trained from scratch with domain randomization (DR) and without.}
        \label{dr-vs-ndr-sr}
\end{figure}

\subsubsection{Domain randomization improves real robot performance}
The performance of agents on the real robot significantly improved after tuning using DR (see Table \ref{dr-vs-ndr}), the detailed experimental results can be seen in Appendix \ref{eval-raw-data-dr}, \ref{eval-raw-data-ndr} and \ref{eval-raw-data-dr-tune}. Among these agents, the agent trained from Seed 123 achieved simulator-level performance. Nonetheless, using DR from the beginning of training is not recommended (at least for our algorithm). In a domain-randomized simulation environment, the agent will be hindered in learning the optimal policy, compared to the policy learned in the standard simulation environment that does not use domain randomization (see Figure \ref{dr-vs-ndr-sr}), and hence will achieve poor performance on the real robots (see Table \ref{dr-vs-ndr}). A demonstration video of this trained network in action can be seen in \url{https://www.youtube.com/watch?v=0Lpod542T9k}

\begin{table}[]
\centering
\caption{Evaluation rewards of different agents evaluated in simulation and on real robots; three seeds used in this paper. Scratch(DR) represents the agent trained from scratch using DR from the beginning of training. Scratch(NDR) is the agent trained from scratch without using DR; Scratch(NDR) + Tune(DR) is the agent firstly trained from scratch without DR and then subsequently tuned using DR in a second training phase. The raw results before averaging are shown in Appendix \ref{eval-raw-data-dr}, \ref{eval-raw-data-ndr} and \ref{eval-raw-data-dr-tune}.}
  \label{dr-vs-ndr}
\begin{tabular}{cccc}
\hline
\textbf{Seed}                            & \textbf{Type}                    & \textbf{Simulation}  & \textbf{Real robot}   \\ \hline
\multirow{3}{*}{\textbf{0}}              & Scratch(DR)                      & -8808                & -16685                \\
                                         & Scratch(NDR)                     & -5447                & -11129                \\
                                         & \textbf{Scratch(NDR) + Tune(DR)} & -5825                & -8922                 \\ \hline
\multirow{3}{*}{\textbf{123}}            & Scratch(DR)                      & -9873                & -15726                \\
                                         & Scratch(NDR)                     & -6376                & -9417                 \\
                                         & \textbf{Scratch(NDR) + Tune(DR)} & -6253                & \textbf{-7030}        \\ \hline
\multirow{3}{*}{\textbf{200}}            & Scratch(DR)                      & -11779               & -15398                \\
                                         & Scratch(NDR)                     & -7277                & -15045                \\
                                         & \textbf{Scratch(NDR) + Tune(DR)} & -6147                & -8535                 \\ \hline
\multirow{3}{*}{\textbf{Average over seeds}} & Scratch(DR)                      & -9913 ± 1866         & -15936 ± 749          \\
                                         & Scratch(NDR)                     & -6367 ± 920          & -11864 ± 3181         \\
                                         & \textbf{Scratch(NDR) + Tune(DR)} & \textbf{-6075 ± 250} & \textbf{-8162 ± 1132} \\ \hline
\end{tabular}
\end{table}

\begin{figure}[hbp]
     \centering
     \begin{subfigure}[b]{0.48\textwidth}
         \centering
         \includegraphics[width=\textwidth]{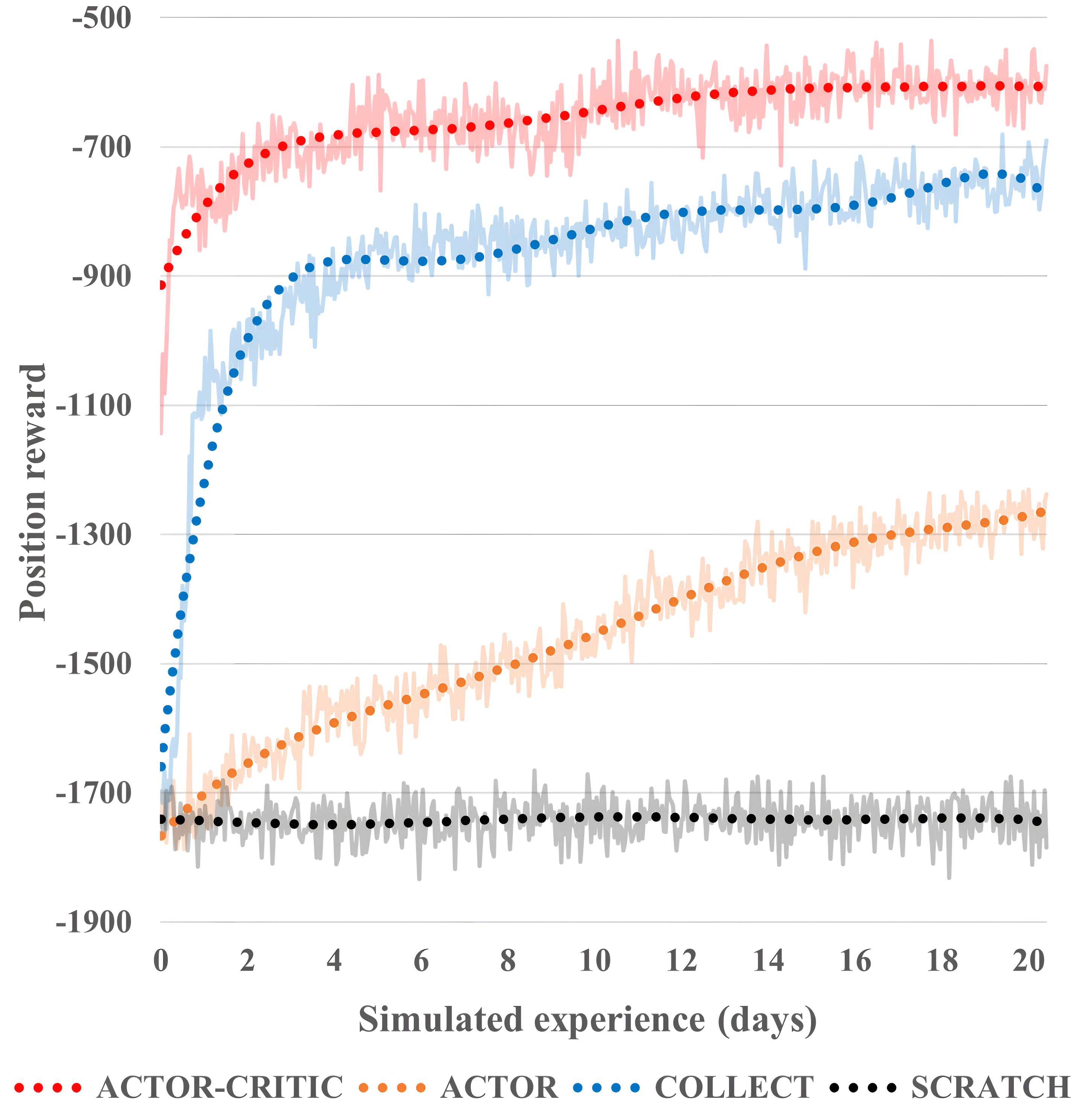}
         \captionsetup{font=footnotesize}
         \subcaption{Position reward vs training time.}
     \end{subfigure}
     \hfill
     \begin{subfigure}[b]{0.48\textwidth}
         \centering
         \includegraphics[width=\textwidth]{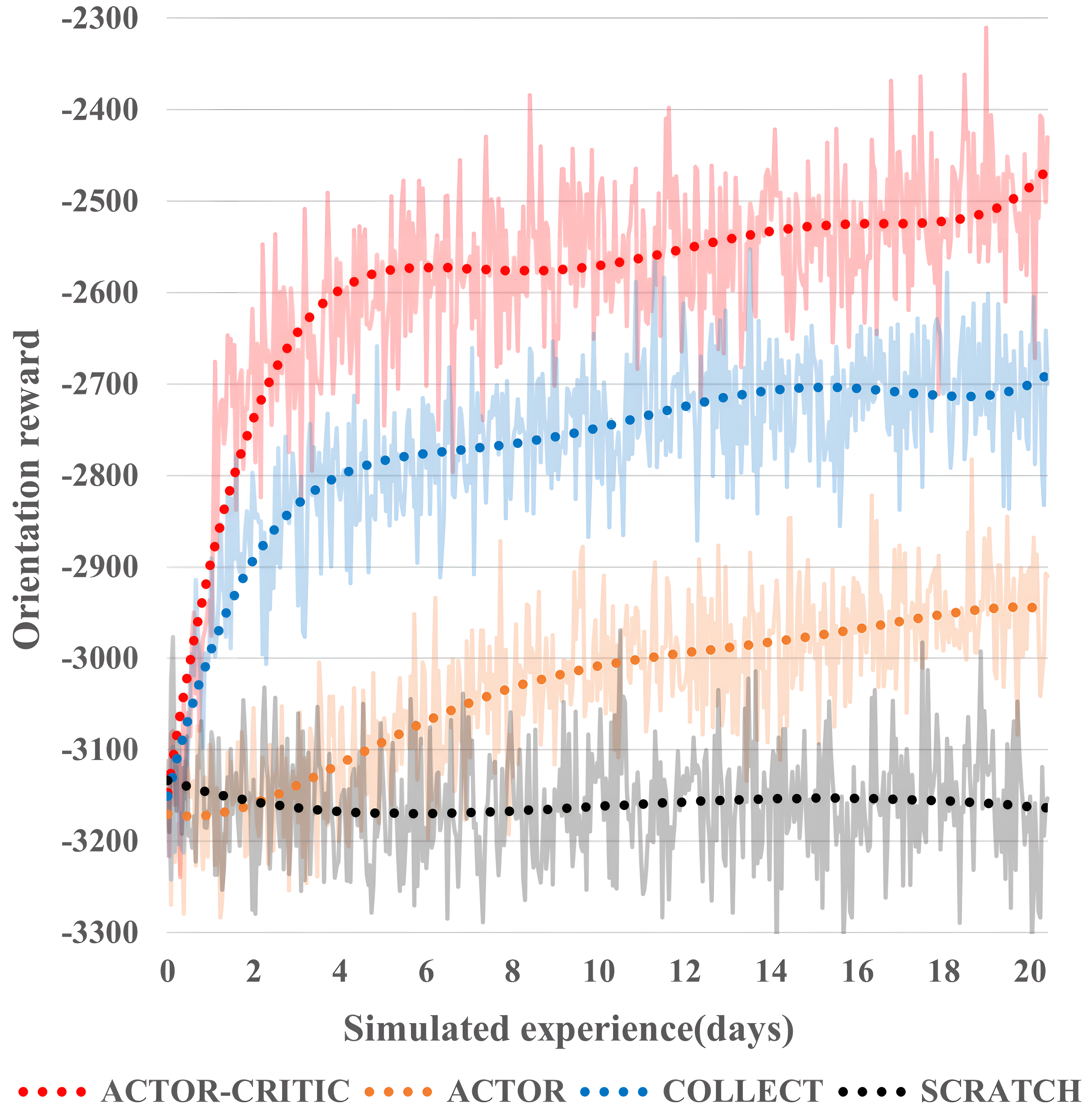}
         \captionsetup{font=footnotesize}
         \subcaption{Orientation reward vs training time.}
     \end{subfigure} 
     \par\bigskip
     \begin{subfigure}[b]{0.48\textwidth}
         \centering
         \includegraphics[width=\textwidth]{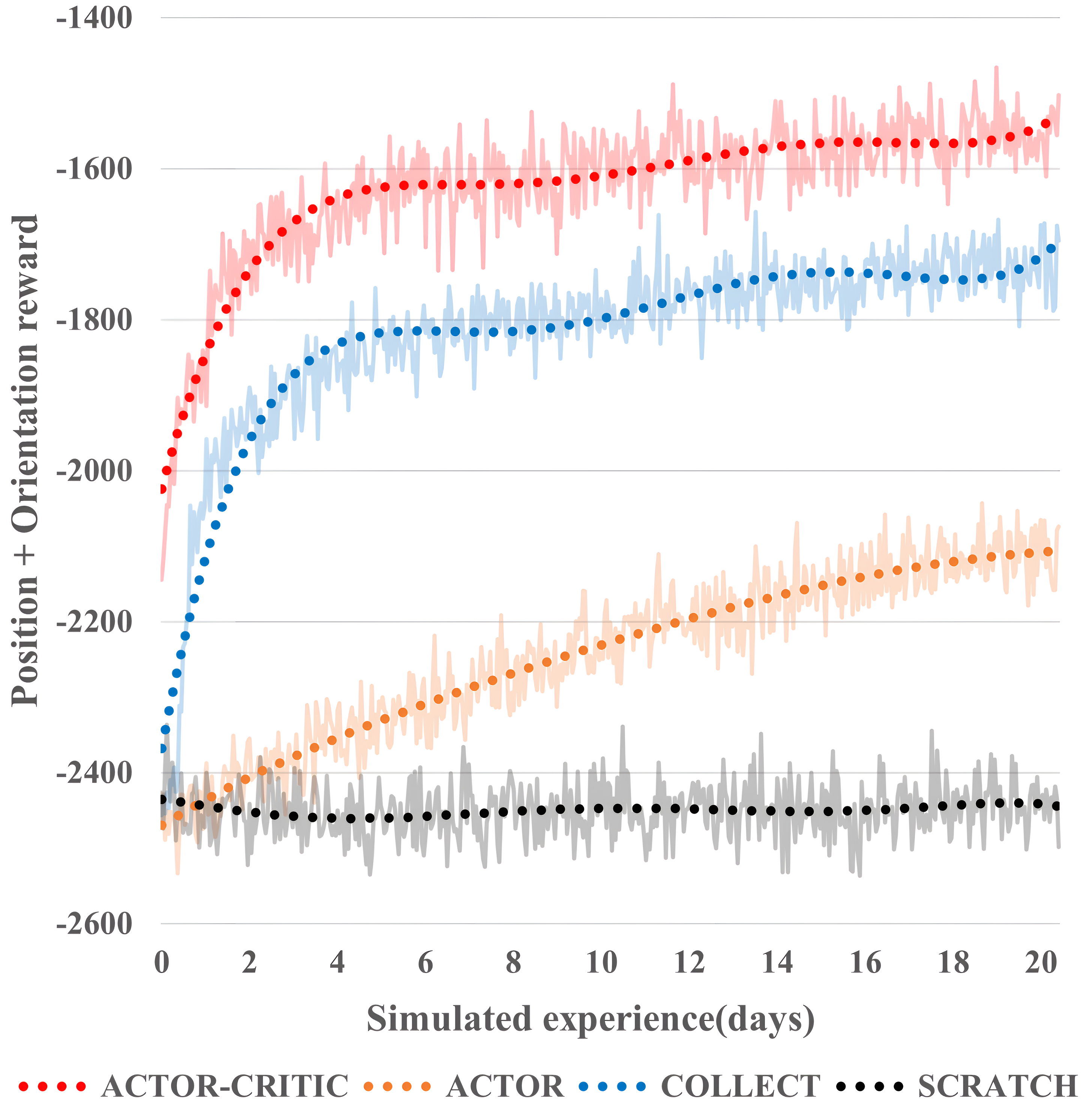}
         \captionsetup{font=footnotesize}
         \subcaption{Combined position and orientation (i.e., pose) reward.}
     \end{subfigure}
        \caption{Learning curves showing rewards achieved by agents using different learning algorithms.}
        \label{st performance}
\end{figure}

\subsection{Move Cube on Trajectory Pro}
Figure \ref{st performance} compares the training performance of different approaches to learning the \textit{Move Cube on Trajectory Pro} task, including using our proposed KT scheme compared to not using it. Agents' overall rewards are decomposed into position and orientation components so that the learning outcomes of the two components can be compared.

Training using the combined DDPG and HER algorithm (not using KT, see \textit{SCRATCH} curve in Figure \ref{st performance}) completely fails in the \textit{Move Cube on Trajectory Pro} task, with rewards on all components staying at their small values. In contrast, agents in all three KT methods shows learning trends. Among them, both \textit{COLLECT} and the \textit{SCRATCH} use randomly initialized neural networks. However, with the help of the experience collected by the teacher, \textit{COLLECT} shows a more successful learning performance.

By transferring the \textit{teacher's} knowledge of how to perform the position-only task to the \textit{ACTOR-CRITIC} \textit{student}, the \textit{ACTOR-CRITIC} excels in its ability to reach position targets from the beginning of training. This makes it more likely that the \textit{ACTOR-CRITIC} will discover actions which change the cube's orientation and are rewarded. All agents learn orientation from scratch, while the \textit{ACTOR-CRITIC} has a better learning performance than that of \textit{COLLECT}. However, the \textit{COLLECT} agent relies strongly on the experience collected by the \textit{teacher} in the early stages of training, although we set a decay on the \textit{teacher's} participation over time. 
Although the \textit{ACTOR} agent inherits the knowledge of the \textit{teacher's} actor, it performs the worst among all KT strategies. Because the randomly-initialized critic is not compatible with the well-trained actor, and the critic's feedback cannot meet the actor's expectations; hence, the actor's performance worsens.

The evaluation results are shown in Table \ref{average deviation} and the raw results before averaging are shown in Appendix \ref{raw-deviation}. All three methods enable the student agent to learn the \textit{Move Cube on Trajectory Pro} task, but behave differently. Among them, the \textit{ACTOR-CRITIC} agent can achieve comparable performance to the \textit{teacher} at reaching position targets, and learn to perform very well on reaching orientation targets. A demonstration video of this trained network in action can be seen here: \url{https://www.youtube.com/watch?v=GhkCqoMqxU4}.

\begin{table}[]
\centering
\caption{The average deviations between achieved and desired position/orientation over 15 evaluation episodes in the simulator; each episode lasts 120,000 time-steps. The deviations are self-customized to see the learning outcomes intuitively. For the position: $\textbf{dev}_\textbf{dis} = ||{g'}_{xyz} - {g}_{xyz}||$, which can measure the distance between the actual position of the cube and the goal position. For the orientation: $\textbf{dev}_\textbf{angle} = {||(R(g'_{o}))^{-1} R(g_{o})||}$, which can measure the angle difference between the actual and target orientations. The deviations of all time-steps are accumulated and finally averaged. The raw results before averaging are shown in Appendix \ref{raw-deviation}.}
  \label{average deviation}
\begin{tabular}{lrrrrr}
\hline
                    & \textbf{ACTOR-CRITIC}  & \textbf{ACTOR} & \textbf{COLLECT} & \textbf{SCRATCH} & \textbf{TEACHER} \\ \hline
\textbf{Average position deviation $\textbf{dev}_\textbf{dis}$ (m)} & \textit{\textbf{0.023}} & 0.066          & 0.031            & 0.134            & 0.024            \\
\textbf{Average orientation deviation $\textbf{dev}_\textbf{angle}$ ($^\circ$)} & \textit{\textbf{75.8}}  & 98.6           & 84.9             & 142.2            & 126.2            \\ \hline
\end{tabular}
\end{table}

\section{Discussion} \label{discussion}
Deep reinforcement learning shows promise for solving robotic manipulation tasks requiring dexterity. Compared with traditional robot control methods, it requires less bespoke engineering work and often has better performance, as shown in our results. We have only tried our approach using the DDPG learning algorithm; other later published DRL algorithms, such as SAC \cite{sac} or TD3 \cite{td3} may have superior performance, and will be the focus of our future research.

Although DRL approaches for solving robotic manipulation tasks are outstanding, the algorithm is somewhat training and sample inefficient. For example, solving the \textit{Move Cube on Trajectory} task takes roughly 10 million environment steps to converge (equivalent to 6 days of simulated experience), while a large part of the time is spent in blind exploration of the environment. The KT method we developed can reduce blind exploration in the initial stage and, more importantly, avoid repeated learning of the same knowledge in different experimental trials. It can improve resource utilization efficiency. A future similar, but innovative research direction would be to train a master \textit{teacher} with broad knowledge in complex environments and use it widely in various downstream tasks\cite{worldknowledge}. Another important direction for future work would be to increase sample efficiency, perhaps achievable via a model-based reinforcement learning approach\cite{mbpo,I-HER}.

The reward function is the driver of DRL; a well-designed and well-shaped reward can improve the quality and speed of DRL \cite{gpu-paper}. A bespoke reward function might limit the utility of a method to a small range of tasks, and is likely costly to develop. In contrast, goal-based sparse rewards can be easily created. Compared to shaped rewards, goal-based sparse rewards have an extremely weak reward signal. The agent can only get the a reward when succeeding in the task; otherwise, it gets punished. However, the probability of success in some tasks is very low, and a dearth of positive feedback hinders improvement in the task. HER can effectively enhance sparse reward signals by modifying the experience data before giving it to the agent. Typically, HER replaces the goal with the posterior state, making the prior-posterior transition highly likely to succeed by the action applied to the prior state. HER has been proven to work wonders \cite{her} in sparse-reward-based tasks, as verified in our case. In addition, our distance-based dense reward can effectively accelerate the agent's learning in the early stages of training, and its implementation is straightforward. We have only validated the approach of mixing sparse and dense rewards for this specific case so far. Indeed, we know from past experience that this method is not required for some "lifting" tasks, such as OpenAI's\cite{open-ai} FetchPickAndPlace-v0, in which the robot has a two-finger gripper that can lift the object relatively easily. However, lifting the object in the $z$-direction using the RRC’s TriFinger robot is much more difficult, requiring dexterous cooperation between three fingers. Therefore, we enhance the $z$-direction reward signal to accelerate the initial stages of training when random exploration is unlikely to lift the object. However, from our experience we can provide the anecdote that excessive use of dense rewards can diminish the benefit of the HER algorithm, which is designed to work with sparse rewards. Hence our method has an appropriate mix of sparse and dense rewards which balances the benefits of using sparse rewards with HER against having dense rewards to acquire difficult skills earlier in learning. We think the principle of our approach can be generalised, but it is not yet clear how to engineer a reward function with the optimal mix of sparse and dense rewards in general.

Sim-to-real transfer is an essential means of bringing robotic learning into reality. Training an intelligent robotic controller in the real world is expensive and impractical. Take an example of a problem we faced before: we tried to collect the data from the real TriFinger robots to train an agent from scratch; however, a large number of interactions caused a lot of friction between the finger and the cube, resulting in wear and tear, which adhered to the surface of the cube and interfered with the estimation of the cube's pose by machine vision algorithms. Indeed, the main limitation of our approach was the absence of any real-robot data. It is likely that some fine-tuning of the policy on real data would greatly increase its robustness in the real environment, and developing a technique which could do so efficiently is one direction for future work. 

Different from the training of a traditional deep learning model, the training of a deep reinforcement learning model requires a large amount of random exploration. Pseudorandom number sequences are generated to drive this (pseudo)random exploration. However, if the same seed is provided to the pseudorandom number generator for the same experiment, the same sequence of exploratory actions will be performed; in this sense the outcome of the reinforcement learning activity is deterministic. The network learned may always converge on the same solution, irrespective of the initial seed chosen. But empirically this is not the case, and the choice of initial seed does influence the performance of the network that is learned. Changing the seed provided to the pseudorandom number generator would allow the sensitivity of the solution to this initial seed to be evaluated; however, this is obviously a very time-consuming process. In this paper, we evaluate three candidates seeds (arbitrarily, integers 0, 123, and 200) to give an indication of the sensitivity of the training process to seed selection. The results show that the reward score achieved varied by ±27\% of the mean across these three seeds. Hence, this shows that when deploying the DRL controller on a robot, the best-performing agent might be selected from a range of policies learned using different seed values.

The development of DRL-driven robotic manipulation is still an unsolved problem, but is currently receiving a lot of attention from the research community. Solving this problem will have a huge impact on industry and society in general. At present, the training of DRL agents is mostly carried out in the simulator, using perfect feedback data which is mostly vision-based. The current capabilities of simulators to render realistic mechanics, including the friction, texture, and deformations at contact interfaces between robotic fingers and the objects they manipulate, are very basic. However, as such simulations improve, and the ability of real world robots to sense these same contact mechanics advances in tandem, we can expect significant improvements in the dexterity of DRL-based robotic manipulation systems. This ability to dexterously interact with and learn from the world is surely the path to developing systems with superior intelligence.

\section{Conclusion} \label{conclusion}
Our relatively simple reinforcement learning approach fully solves the \textit{Move Cube on Trajectory} task in simulation. Unlike the RRC 2020 benchmark solution \cite{dr-code}, this was achieved with the use of minimal domain-specific knowledge. Reproducing our method is highly feasible, since only one machine with eight processors is required; no GPUs are required. The number of agents can be set to more if an increase in training speed is desirable. Or, the number of agents can be set to less, but sacrificing training speed. 

Moreover, our learned policies can be successfully deployed on the real robot through our DR tuning approach, which bridges the large sim-to-real domain gap and allows the achievement of near-simulator-level performance on real robots. In the RRC 2021, which used the \textit{Move Cube on Trajectory} task, requiring only position targets be reached, we outperformed all competing submissions, including those employing more classical robotic control techniques. 

Our novel KT approach can enable an agent to perform more useful interactions with the environment in the early stages of learning, helping to avoid exploration actions that do not result in learning experiences. It allows the complex \textit{Move Cube on Trajectory Pro} task to be solved efficiently in the simulator. Additionally, KT increases resource utilization efficiency, avoiding repeatedly re-learning similar skills. It may be feasible to deploy this method in any actor-critic RL algorithm, and the implementation method is straightforward. 

\section*{Acknowledgments}
This publication has emanated from research conducted with the financial support of Science Foundation Ireland under grant numbers 17/FRL/4832 and SFI/12/RC/2289\_P2 and of China Scholarship Council (CSC No.202006540003). For the purpose of Open Access, the author has applied a CC BY public copyright licence to any Author Accepted Manuscript version arising from this submission. We acknowledge the Research IT HPC Service at University College Dublin for providing computational facilities and support that contributed to the research results reported in this paper.

\setlength{\bibsep}{0em}

\newpage
\appendix
\section{Neural network} \label{neural-network-params}
\subsection{Architecture}
\begin{figure}[]
     \centering
     \begin{subfigure}[b]{0.43\textwidth}
         \centering
         \includegraphics[width=\textwidth]{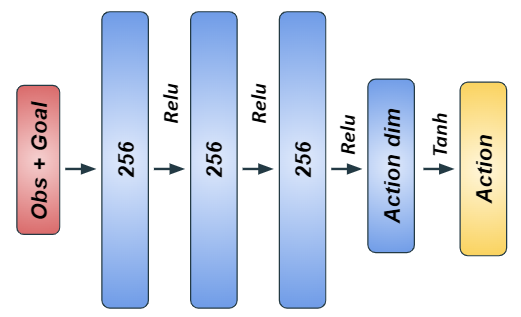}
         \captionsetup{font=footnotesize}
         \subcaption{The structure of actor neural network.}
     \end{subfigure}
     \hfill
     \begin{subfigure}[b]{0.38\textwidth}
         \centering
         \includegraphics[width=\textwidth]{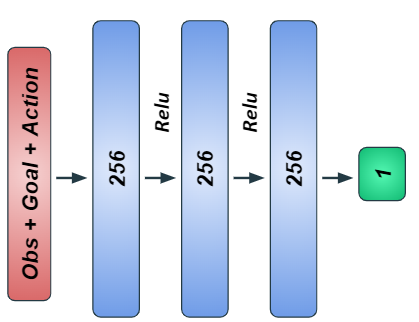}
         \captionsetup{font=footnotesize}
         \subcaption{The structure of critic neural network.}
     \end{subfigure}
        \caption{The neural network structure of the actor and critic. The action output by the actor multiplies the max action ratio before applying it to the environment to make sure it cannot exceed the limits of the action space.}
        \label{fig:actor-critic of ddpg}
\end{figure}

The code used in our work is mostly from \url{https://github.com/TianhongDai/hindsight-experience-replay.}, and we have made some edits to adapt it to our environment. The code of \textit{Move Cube on Trajectory} can be found here \url{https://github.com/RobertMcCarthy97/rrc_phase1}, and the code of \textit{Move Cube on Trajactary Pro} can be found here \url{https://github.com/wq13552463699/rrc_pos_ori}.
The structures of the actor and critic neural networks are shown in Figure \ref{fig:actor-critic of ddpg}. 

\subsection{Hyper-parameter}
Hyperparameter related to the task/training/testing can be seen in the sub-folder $rrc\_example\_package/her/arguments.py$ of the GitHub repositories above.
\begin{itemize}
    \item \textit{Learning rate of the actor}: 0.001
    \item \textit{Learning rate of the critic}: 0.001
    \item \textit{Discount factor}: 0.98
    \item \textit{Batch size}: 256
    \item \textit{The average coefficient for soft-update of the target networks}: 0.95
    \item \textit{Replay buffer size}: 1,000,000
    \item \textit{HER replay strategy}: \textit{future}
\end{itemize}



\newpage
\section{Raw data}
\begin{table*}[hdp]
\centering
\caption{The evaluated scores of DRL agents trained with different seeds on both simulator and real robot. All DRL agents are trained with DR from scratch. We did not run multiple times for each agent on the real robot, because agent performances are generally poor. This table is the a part of the raw data of Table \ref{dr-vs-ndr}, before averaging.}
\scalebox{0.9}{
\begin{tabular}{|c|cc|cc|cc|}
               & \multicolumn{2}{c|}{\textbf{Seed 0}} & \multicolumn{2}{c|}{\textbf{Seed 123}} & \multicolumn{2}{c|}{\textbf{Seed 200}} \\
\textbf{Num} &
  \multicolumn{1}{c|}{\textbf{Rob1}} &
  \textbf{Sim} &
  \multicolumn{1}{c|}{\textbf{Rob6}} &
  \textbf{Sim} &
  \multicolumn{1}{c|}{\textbf{Rob5}} &
  \textbf{Sim} \\
\textbf{1}  & \multicolumn{1}{c|}{-12115} & -8288  & \multicolumn{1}{c|}{-15639}  & -9200   & \multicolumn{1}{c|}{-10120}  & -8379   \\
\textbf{2}  & \multicolumn{1}{c|}{-15167} & -9430  & \multicolumn{1}{c|}{-14262}  & -12781  & \multicolumn{1}{c|}{-16159}  & -12284  \\
\textbf{3}  & \multicolumn{1}{c|}{-23378} & -8290  & \multicolumn{1}{c|}{-8257}   & -10560  & \multicolumn{1}{c|}{-21487}  & -9197   \\
\textbf{4}  & \multicolumn{1}{c|}{-21338} & -10045 & \multicolumn{1}{c|}{-11242}  & -9759   & \multicolumn{1}{c|}{-23810}  & -7111   \\
\textbf{5}  & \multicolumn{1}{c|}{-25971} & -7998  & \multicolumn{1}{c|}{-11947}  & -8762   & \multicolumn{1}{c|}{-10207}  & -9745   \\
\textbf{6}  & \multicolumn{1}{c|}{-13497} & -8878  & \multicolumn{1}{c|}{-18063}  & -8536   & \multicolumn{1}{c|}{-12042}  & -9183   \\
\textbf{7}  & \multicolumn{1}{c|}{-17905} & -9567  & \multicolumn{1}{c|}{-13341}  & -10685  & \multicolumn{1}{c|}{-12958}  & -9670   \\
\textbf{8}  & \multicolumn{1}{c|}{-22253} & -9332  & \multicolumn{1}{c|}{-15494}  & -10022  & \multicolumn{1}{c|}{-15986}  & -15520  \\
\textbf{9}  & \multicolumn{1}{c|}{-11608} & -7669  & \multicolumn{1}{c|}{-12824}  & -7113   & \multicolumn{1}{c|}{-18867}  & -10239  \\
\textbf{10} & \multicolumn{1}{c|}{-11805} & -8351  & \multicolumn{1}{c|}{-21261}  & -15298  & \multicolumn{1}{c|}{-17593}  & -8557   \\
\textbf{11} & \multicolumn{1}{c|}{-10500} & -8677  & \multicolumn{1}{c|}{-20369}  & -9053   & \multicolumn{1}{c|}{-16410}  & -16587  \\
\textbf{12} & \multicolumn{1}{c|}{-19996} & -9367  & \multicolumn{1}{c|}{-23302}  & -9123   & \multicolumn{1}{c|}{-18469}  & -11242  \\
\textbf{13} & \multicolumn{1}{c|}{-20676} & -8585  & \multicolumn{1}{c|}{-18059}  & -8727   & \multicolumn{1}{c|}{-11228}  & -13206  \\
\textbf{14} & \multicolumn{1}{c|}{-12666} & -7764  & \multicolumn{1}{c|}{-18087}  & -8337   & \multicolumn{1}{c|}{-10918}  & -12380  \\
\textbf{15} & \multicolumn{1}{c|}{-11413} & -9877  & \multicolumn{1}{c|}{-13749}  & -10143  & \multicolumn{1}{c|}{-14727}  & -23389  \\
\textit{\textbf{Avg}} &
  \multicolumn{1}{c|}{\textit{\textbf{-16686}}} &
  \textit{\textbf{-8808}} &
  \multicolumn{1}{c|}{\textit{\textbf{-15726}}} &
  \textit{\textbf{-9873}} &
  \multicolumn{1}{c|}{\textit{\textbf{-15399}}} &
  \textit{\textbf{-11779}}
\end{tabular}
}
\label{eval-raw-data-dr}
\end{table*}

\begin{table*}[hdp]
\centering
\caption{The evaluated scores of DRL agents trained with different seeds on both simulator and real-robot. All DRL agents are trained from scratch without DR. This table is the raw data of Table \ref{dr-vs-ndr} and Table \ref{real and sim eval different seeds} before averaging.}
\scalebox{0.9}{
\begin{tabular}{|c|cccc|cccc|cccc|}
 &
  \multicolumn{4}{c|}{\textbf{Seed 0}} &
  \multicolumn{4}{c|}{\textbf{Seed 123}} &
  \multicolumn{4}{c|}{\textbf{Seed 200}} \\
\textbf{Num} &
  \multicolumn{1}{l}{\textbf{Rob1}} &
  \multicolumn{1}{l}{\textbf{Rob5}} &
  \multicolumn{1}{l|}{\textbf{Robo6}} &
  \multicolumn{1}{l|}{\textbf{Sim}} &
  \textbf{Rob1} &
  \textbf{Rob5} &
  \multicolumn{1}{c|}{\textbf{Rob6}} &
  \multicolumn{1}{l|}{\textbf{Sim}} &
  \multicolumn{1}{c}{\textbf{Rob1}} &
  \textbf{Rob5} &
  \multicolumn{1}{c|}{\textbf{Rob6}} &
  \multicolumn{1}{l|}{\textbf{Sim}} \\
\textbf{1} &
  -9970 &
  -11190 &
  \multicolumn{1}{l|}{-11512} &
  -5635 &
  -8789 &
  -7328 &
  \multicolumn{1}{r|}{-10030} &
  -6579 &
  -22824 &
  -12436 &
  \multicolumn{1}{r|}{-11927} &
  -6965 \\
\textbf{2} &
  -11412 &
  -9017 &
  \multicolumn{1}{l|}{-11461} &
  -5029 &
  -8086 &
  -10263 &
  \multicolumn{1}{r|}{-13828} &
  -6650 &
  -12288 &
  -10107 &
  \multicolumn{1}{r|}{-19633} &
  -7464 \\
\textbf{3} &
  -10746 &
  -12084 &
  \multicolumn{1}{l|}{-10900} &
  -5175 &
  -11328 &
  -9908 &
  \multicolumn{1}{r|}{-8270} &
  -7180 &
  -10859 &
  -9501 &
  \multicolumn{1}{r|}{-17123} &
  -8839 \\
\textbf{4} &
  -7572 &
  -15767 &
  \multicolumn{1}{l|}{-11904} &
  -5570 &
  -9928 &
  -8228 &
  \multicolumn{1}{r|}{-9110} &
  -6314 &
  -19834 &
  -11469 &
  \multicolumn{1}{r|}{-15823} &
  -5422 \\
\textbf{5} &
  -10883 &
  -10430 &
  \multicolumn{1}{l|}{-13682} &
  -5430 &
  -13106 &
  -8263 &
  \multicolumn{1}{r|}{-10662} &
  -5746 &
  -13174 &
  -17859 &
  \multicolumn{1}{r|}{-9643} &
  -7505 \\
\textbf{6} &
  -7896 &
  -11427 &
  \multicolumn{1}{l|}{-10371} &
  -5895 &
  -11644 &
  -10245 &
  \multicolumn{1}{r|}{-6619} &
  -5932 &
  -18568 &
  -14007 &
  \multicolumn{1}{r|}{-17594} &
  -7324 \\
\textbf{7} &
  -7942 &
  -14505 &
  \multicolumn{1}{l|}{-14028} &
  -4977 &
  -6447 &
  -11307 &
  \multicolumn{1}{r|}{-12355} &
  -6106 &
  -12754 &
  -10631 &
  \multicolumn{1}{r|}{-11542} &
  -6473 \\
\textbf{8} &
  -8461 &
  -11135 &
  \multicolumn{1}{l|}{-13604} &
  -5698 &
  -9180 &
  -7481 &
  \multicolumn{1}{r|}{-11375} &
  -6551 &
  -21177 &
  -11721 &
  \multicolumn{1}{r|}{-14255} &
  -6009 \\
\textbf{9} &
  -6441 &
  -12871 &
  \multicolumn{1}{l|}{-9460} &
  -4921 &
  -10841 &
  -6781 &
  \multicolumn{1}{r|}{-7524} &
  -6560 &
  -12986 &
  -11987 &
  \multicolumn{1}{r|}{-15043} &
  -7716 \\
\textbf{10} &
  -11466 &
  -11620 &
  \multicolumn{1}{l|}{-10030} &
  -5702 &
  -4951 &
  -8415 &
  \multicolumn{1}{r|}{-10385} &
  -6257 &
  -10326 &
  -12408 &
  \multicolumn{1}{r|}{-17208} &
  -8291 \\
\textbf{11} &
  -9513 &
  -5192 &
  \multicolumn{1}{l|}{-15004} &
  -5259 &
  -9020 &
  -8248 &
  \multicolumn{1}{r|}{-7751} &
  -6055 &
  -18069 &
  -11129 &
  \multicolumn{1}{r|}{-19162} &
  -7405 \\
\textbf{12} &
  -11928 &
  -10046 &
  \multicolumn{1}{l|}{-11051} &
  -6016 &
  -6250 &
  -9530 &
  \multicolumn{1}{r|}{-11040} &
  -6722 &
  -19168 &
  -19332 &
  \multicolumn{1}{r|}{-18368} &
  -6590 \\
\textbf{13} &
  -14798 &
  -11213 &
  \multicolumn{1}{l|}{-12311} &
  -5027 &
  -8987 &
  -10205 &
  \multicolumn{1}{r|}{-11442} &
  -7148 &
  -15011 &
  -15671 &
  \multicolumn{1}{r|}{-15546} &
  -8015 \\
\textbf{14} &
  -10501 &
  -12419 &
  \multicolumn{1}{l|}{-16049} &
  -5531 &
  -8834 &
  -9002 &
  \multicolumn{1}{r|}{-9852} &
  -6101 &
  -23270 &
  -17366 &
  \multicolumn{1}{r|}{-19554} &
  -6483 \\
\textbf{15} &
  -9178 &
  -10589 &
  \multicolumn{1}{l|}{-11256} &
  -5847 &
  -12243 &
  -8691 &
  \multicolumn{1}{r|}{-16036} &
  -5732 &
  -13678 &
  -11187 &
  \multicolumn{1}{r|}{-13829} &
  -8657 \\
\textit{\textbf{Avg}} &
  \textit{\textbf{-9914}} &
  \textit{\textbf{-11300}} &
  \multicolumn{1}{c|}{\textit{\textbf{-12175}}} &
  \textit{\textbf{-5448}} &
  \textit{\textbf{-9309}} &
  \textit{\textbf{-8926}} &
  \multicolumn{1}{r|}{\textit{\textbf{-10017}}} &
  \textit{\textbf{-6376}} &
  \textit{\textbf{-16266}} &
  \textit{\textbf{-13121}} &
  \multicolumn{1}{r|}{\textit{\textbf{-15750}}} &
  \textit{\textbf{-7277}}
\end{tabular}
}
\label{eval-raw-data-ndr}
\end{table*}

\begin{table*}[h]
\centering
\caption{The evaluated scores of DRL agents trained with different seeds on both simulator and real-robot. All DRL agents are trained from scratch without DR and then tuned by DR. This table is a part of the raw data of Table \ref{dr-vs-ndr} before averaging.}
\scalebox{0.9}{
\begin{tabular}{|c|cccc|cccc|cccc|}
 &
  \multicolumn{4}{c|}{\textbf{Seed 0}} &
  \multicolumn{4}{c|}{\textbf{Seed 123}} &
  \multicolumn{4}{c|}{\textbf{Seed 200}} \\
\textbf{Num} &
  \multicolumn{1}{l}{\textbf{Rob1}} &
  \multicolumn{1}{l}{\textbf{Rob5}} &
  \multicolumn{1}{l|}{\textbf{Rob6}} &
  \textbf{Sim} &
  \multicolumn{1}{c}{\textbf{Rob1}} &
  \multicolumn{1}{c}{\textbf{Rob5}} &
  \multicolumn{1}{c|}{\textbf{Rob6}} &
  \textbf{Sim} &
  \multicolumn{1}{c}{\textbf{Rob1}} &
  \multicolumn{1}{c}{\textbf{Rob5}} &
  \multicolumn{1}{c|}{\textbf{Rob6}} &
  \textbf{Sim} \\
\textbf{1} &
  -8787 &
  -9647 &
  \multicolumn{1}{r|}{-7396} &
  -5257 &
  -7579 &
  -6335 &
  \multicolumn{1}{r|}{-6700} &
  -6461 &
  -8000 &
  -10161 &
  \multicolumn{1}{r|}{-7614} &
  -5413 \\
\textbf{2} &
  -11681 &
  -6962 &
  \multicolumn{1}{r|}{-10088} &
  -6809 &
  -6344 &
  -6729 &
  \multicolumn{1}{r|}{-7252} &
  -5724 &
  -8219 &
  -9995 &
  \multicolumn{1}{r|}{-8555} &
  -6481 \\
\textbf{3} &
  -6013 &
  -8187 &
  \multicolumn{1}{r|}{-6318} &
  -5442 &
  -7376 &
  -6124 &
  \multicolumn{1}{r|}{-5388} &
  -7291 &
  -13052 &
  -7443 &
  \multicolumn{1}{r|}{-6707} &
  -6812 \\
\textbf{4} &
  -10352 &
  -6727 &
  \multicolumn{1}{r|}{-4912} &
  -5169 &
  -5364 &
  -8417 &
  \multicolumn{1}{r|}{-7932} &
  -6066 &
  -12318 &
  -8324 &
  \multicolumn{1}{r|}{-7532} &
  -7491 \\
\textbf{5} &
  -8789 &
  -7936 &
  \multicolumn{1}{r|}{-7370} &
  -6081 &
  -9498 &
  -8711 &
  \multicolumn{1}{r|}{-6013} &
  -6474 &
  -10980 &
  -7155 &
  \multicolumn{1}{r|}{-8574} &
  -5620 \\
\textbf{6} &
  -10460 &
  -11045 &
  \multicolumn{1}{r|}{-7351} &
  -5908 &
  -8529 &
  -7367 &
  \multicolumn{1}{r|}{-5514} &
  -5788 &
  -11110 &
  -6375 &
  \multicolumn{1}{r|}{-7533} &
  -8658 \\
\textbf{7} &
  -8279 &
  -10296 &
  \multicolumn{1}{r|}{-11356} &
  -6426 &
  -6393 &
  -4906 &
  \multicolumn{1}{r|}{-8937} &
  -6458 &
  -7776 &
  -6660 &
  \multicolumn{1}{r|}{-7679} &
  -5728 \\
\textbf{8} &
  -10261 &
  -9046 &
  \multicolumn{1}{r|}{-6375} &
  -6043 &
  -4557 &
  -7492 &
  \multicolumn{1}{r|}{-7335} &
  -5680 &
  -8995 &
  -9554 &
  \multicolumn{1}{r|}{-6193} &
  -6664 \\
\textbf{9} &
  -6371 &
  -11219 &
  \multicolumn{1}{r|}{-6704} &
  -4671 &
  -7729 &
  -6269 &
  \multicolumn{1}{r|}{-7673} &
  -6544 &
  -9650 &
  -7126 &
  \multicolumn{1}{r|}{-7821} &
  -6263 \\
\textbf{10} &
  -10443 &
  -11136 &
  \multicolumn{1}{r|}{-11567} &
  -6491 &
  -8103 &
  -9095 &
  \multicolumn{1}{r|}{-9221} &
  -7921 &
  -10734 &
  -9523 &
  \multicolumn{1}{r|}{-5636} &
  -6289 \\
\textbf{11} &
  -6236 &
  -7585 &
  \multicolumn{1}{r|}{-11795} &
  -5751 &
  -5496 &
  -7470 &
  \multicolumn{1}{r|}{-5685} &
  -5299 &
  -11455 &
  -7360 &
  \multicolumn{1}{r|}{-6246} &
  -5608 \\
\textbf{12} &
  -8060 &
  -6528 &
  \multicolumn{1}{r|}{-10957} &
  -3804 &
  -6306 &
  -9059 &
  \multicolumn{1}{r|}{-8037} &
  -5429 &
  -8518 &
  -5751 &
  \multicolumn{1}{r|}{-9680} &
  -4373 \\
\textbf{13} &
  -9808 &
  -8811 &
  \multicolumn{1}{r|}{-8557} &
  -5418 &
  -5114 &
  -6558 &
  \multicolumn{1}{r|}{-8175} &
  -5993 &
  -7719 &
  -9358 &
  \multicolumn{1}{r|}{-9330} &
  -5051 \\
\textbf{14} &
  -10962 &
  -11804 &
  \multicolumn{1}{r|}{-11804} &
  -7607 &
  -6388 &
  -6056 &
  \multicolumn{1}{r|}{-6038} &
  -6394 &
  -10331 &
  -7711 &
  \multicolumn{1}{r|}{-8775} &
  -5891 \\
\textbf{15} &
  -7681 &
  -7249 &
  \multicolumn{1}{r|}{-10592} &
  -6500 &
  -7062 &
  -6800 &
  \multicolumn{1}{r|}{-7225} &
  -6278 &
  -10197 &
  -8043 &
  \multicolumn{1}{r|}{-6610} &
  -5868 \\
\textit{\textbf{Avg}} &
  \textit{\textbf{-8946}} &
  \textit{\textbf{-8945}} &
  \multicolumn{1}{r|}{\textit{\textbf{-8876}}} &
  \textit{\textbf{-5825}} &
  \textit{\textbf{-6789}} &
  \textit{\textbf{-7159}} &
  \multicolumn{1}{r|}{\textit{\textbf{-7142}}} &
  \textit{\textbf{-6253}} &
  \textit{\textbf{-9937}} &
  \textit{\textbf{-8036}} &
  \multicolumn{1}{r|}{\textit{\textbf{-7632}}} &
  \textit{\textbf{-6147}}
\end{tabular}
}
\label{eval-raw-data-dr-tune}
\end{table*}

\begin{table*}[]
\centering
\caption{The evaluated performance of agent with different KT policies. Where Pos is the deviation between actual and desired position and Ori is the deviation of the orientation. The deviation is calculated by averaging the accumulated reward over 120,000 time-step.}
\scalebox{0.9}{
\begin{tabular}{|c|cc|cc|cc|cc|cc|}
 &
  \multicolumn{2}{c|}{\textbf{ACTOR-CRITIC}} &
  \multicolumn{2}{c|}{\textbf{ACTOR}} &
  \multicolumn{2}{c|}{\textbf{COLLECT}} &
  \multicolumn{2}{c|}{\textbf{SCRATCH}} &
  \multicolumn{2}{c|}{\textbf{TEACHER}} \\
\textbf{Num} &
  \textbf{Pos (m)} &
  \textbf{Ori (rad)} &
  \textbf{Pos (m)} &
  \textbf{Ori (rad)} &
  \textbf{Pos (m)} &
  \textbf{Ori (rad)} &
  \textbf{Pos (m)} &
  \textbf{Ori (rad)} &
  \textbf{Pos (m)} &
  \textbf{Ori (rad)} \\
\textbf{1} &
  0.0201 &
  1.3000 &
  0.0698 &
  1.7481 &
  0.0248 &
  1.5535 &
  0.1654 &
  2.5957 &
  0.0215 &
  2.0159 \\
\textbf{2} &
  0.0253 &
  1.2571 &
  0.0643 &
  1.6544 &
  0.0304 &
  1.6287 &
  0.1143 &
  2.4134 &
  0.0243 &
  2.0639 \\
\textbf{3} &
  0.0185 &
  1.4060 &
  0.0579 &
  1.7415 &
  0.0301 &
  1.4512 &
  0.1512 &
  2.3510 &
  0.0178 &
  2.1255 \\
\textbf{4} &
  0.0227 &
  1.3226 &
  0.0797 &
  1.8560 &
  0.0276 &
  1.5742 &
  0.1127 &
  2.5419 &
  0.0304 &
  2.3097 \\
\textbf{5} &
  0.0221 &
  1.2786 &
  0.0651 &
  1.8221 &
  0.0341 &
  1.4510 &
  0.1208 &
  2.2099 &
  0.0242 &
  2.2001 \\
\textbf{6} &
  0.0262 &
  1.3984 &
  0.0664 &
  1.7988 &
  0.0271 &
  1.4512 &
  0.1310 &
  2.3118 &
  0.0274 &
  2.3187 \\
\textbf{7} &
  0.0243 &
  1.2020 &
  0.0691 &
  1.6806 &
  0.0349 &
  1.5679 &
  0.1049 &
  2.4833 &
  0.0204 &
  2.3480 \\
\textbf{8} &
  0.0198 &
  1.1144 &
  0.0619 &
  1.6527 &
  0.0226 &
  1.3843 &
  0.1732 &
  2.3918 &
  0.0277 &
  2.4137 \\
\textbf{9} &
  0.0308 &
  1.3189 &
  0.0639 &
  1.6251 &
  0.0297 &
  1.4060 &
  0.1406 &
  2.5777 &
  0.0326 &
  2.2128 \\
\textbf{10} &
  0.0219 &
  1.2269 &
  0.0611 &
  1.6519 &
  0.0340 &
  1.7080 &
  0.1248 &
  2.6528 &
  0.0264 &
  2.1901 \\
\textbf{11} &
  0.0225 &
  1.7139 &
  0.0601 &
  1.6495 &
  0.0339 &
  1.4148 &
  0.1215 &
  2.5674 &
  0.0226 &
  2.1884 \\
\textbf{12} &
  0.0206 &
  1.5068 &
  0.0738 &
  1.8200 &
  0.0320 &
  1.4506 &
  0.1141 &
  2.4648 &
  0.0238 &
  2.2346 \\
\textbf{13} &
  0.0216 &
  1.3547 &
  0.0694 &
  1.7752 &
  0.0331 &
  1.3885 &
  0.1280 &
  2.4678 &
  0.0200 &
  1.9490 \\
\textbf{14} &
  0.0255 &
  1.2386 &
  0.0626 &
  1.6965 &
  0.0357 &
  1.3612 &
  0.1348 &
  2.5125 &
  0.0242 &
  2.2067 \\
\textbf{15} &
  0.0227 &
  1.1980 &
  0.0666 &
  1.7335 &
  0.0287 &
  1.4212 &
  0.1604 &
  2.6438 &
  0.0179 &
  2.2718 \\
\textbf{Avg} &
  \textit{\textbf{0.0230}} &
  \textit{\textbf{1.3225}} &
  \textit{\textbf{0.0661}} &
  \textit{\textbf{1.7270}} &
  \textit{\textbf{0.0306}} &
  \textit{\textbf{1.4808}} &
  \textit{\textbf{0.1332}} &
  \textit{\textbf{2.4790}} &
  \textit{\textbf{0.0241}} &
  \textit{\textbf{2.2033}}
\end{tabular}
}
\label{raw-deviation}
\end{table*}

\end{document}